\documentclass[10pt, a4paper]{article}

\usepackage[final]{lrec-coling2024} 

\usepackage{amsmath,amssymb,amsfonts}
\usepackage{lipsum} 
\usepackage{float} 
\usepackage{booktabs} 
\usepackage{multirow}
\usepackage{pifont}
\usepackage{hyperref}
\usepackage{graphicx}
\usepackage[]{todonotes}  
\usepackage{fancyhdr}

\newcommand{\cmark}{\ding{51}}
\newcommand{\xmark}{\ding{55}}

\fancypagestyle{firstpagefooter}{
  \fancyhf{}

  \vspace*{-1ex}
  \fancyfoot[C]{%
    \fontfamily{ptm}\selectfont
    \fontsize{11pt}{11pt}\selectfont
    \thepage\\[1.5ex] 
    \fontsize{9pt}{11pt}\selectfont
    \textit{NLPerspectives 2024@LREC-COLING 2024}, pages 100--110\\
    21 May, 2024. © 2024 ELRA Language Resource Association: CC BY-NC 4.0
  }
}

\fancypagestyle{plain}{
  \fancyhf{}

  \fancyfoot[C]{%
    \fontfamily{ptm}\selectfont
    \fontsize{11pt}{11pt}\selectfont
    \thepage
  }
}

\title{The Effectiveness of LLMs as Annotators: A Comparative Overview and Empirical 
Analysis of Direct Representation} 

\name{Maja Pavlovic$^1$ and 
     Massimo Poesio$^{1,2}$ 
     }

\address{$^1$Queen Mary University of London,
         $^2$University of Utrecht\\
         m.pavlovic@qmul.ac.uk,
         m.poesio@\{qmul.ac.uk,uu.nl\}
         }

\abstract{
Large Language Models (LLMs) have emerged as powerful support tools across various natural language tasks and a range of application domains. Recent studies focus on exploring their capabilities for data annotation. 
This paper provides a comparative overview of twelve studies investigating the potential of LLMs in labelling data. While the models demonstrate promising cost and time-saving benefits, there exist considerable limitations, such as representativeness, bias, sensitivity to prompt variations and English language preference. Leveraging insights from these studies, our empirical analysis further examines the alignment between human and GPT-generated opinion distributions across four subjective datasets. In contrast to the studies examining representation, our methodology directly obtains the opinion distribution from GPT. 
Our analysis thereby supports the minority of studies that are considering diverse perspectives when evaluating data annotation tasks and highlights the need for further research in this direction.
\\ \newline  }

\begin{document}
\thispagestyle{firstpagefooter}
\maketitleabstract

\section{Introduction}
Large Language Models (LLMs) have shown impressive abilities in a variety of natural language related tasks \citep{brown2020language, touvron2023llama}.
\citet{brown2020language} demonstrate their ability as few-shot learners and \citet{flan_2021,kojima2022large} evidence their zero-shot capabilities. 
Recognising the significance and costliness of annotated data across various research domains, recent work explores the potential of LLMs as data annotators, encompassing both zero- and few-shot learning approaches \citep{lee2023can, Ziems_Held_Shaikh_Chen_Zhang_Yang_2023, tornbergchatgpt, zhu2023can,gilardi2023chatgpt, mohta2023large, ding2022gpt, he2023annollm}. 
Considering that LLMs are trained to adhere to instructions guided by human preference \citep{ouyang2022training, rafailov2023direct}, studies examine the extent to which human disagreement is captured \citep{lee2023can} and whether or not such disagreement aligns with that of humans \citep{santurkar2023whose}.

Our work, firstly, offers a comparative overview of twelve previous studies that investigate the capabilities of LLMs as annotators, 
concentrating on classification tasks and considering whether disagreement is captured by the studies. 
Secondly, we present an empirical analysis concentrating more specifically on the perspectivist question. We compare the top-performing LLM from the first section (GPT) against human annotators, by examining the degree of alignment 
between their opinion distributions, 
for the case of the four subjective datasets recently used for the 2023 SEMEVAL  Task on Learning With Disagreement \cite{leonardelli_2023_semeval}.

\section{Comparative Overview}
\label{sec:literature_review}
Labelled data forms the foundation for training supervised models across diverse machine learning tasks.
Much recent research has focused on
exploring the use of LLMs as a quicker and more cost-effective alternative to traditional data annotation.
In this first Section we review the research in this area. 
\textcolor{black}{
Due to rapid developments in this space, we concentrate on works from the past year which leverage recent models with a focus on classification tasks. Our approach to selecting relevant papers followed a combination of keyword searches, monitoring relevant workshops and conferences, and examining 
citations.} 

\paragraph{Studies: }
\citet{wang2021want} employ GPT-3 for the annotation of 
datasets, which are subsequently used in the training of smaller models. \citet{huang2023chatgpt} explore the capability of ChatGPT to accurately label implicit hate speech and provide good explanations for its annotations. \citet{zhu2023can} also investigate the capability of GPT for labelling and \citet{he2023annollm} introduce a two step approach in which they first prompt the LLM to generate explanations and then annotate a sample to improve the annotation quality of LLMs. Both \citet{tornbergchatgpt, gilardi2023chatgpt} contrast the performance of GPT with that of crowd-workers. Whereas, \citet{goel2023llms}  introduce a two-stage semi-automated approach employing LLMs and human experts to accelerate annotation for the extraction of medical information. 
\citet{Ziems_Held_Shaikh_Chen_Zhang_Yang_2023} conduct a large scale empirical analysis to understand the zero-shot performance of GPT and Flan on 25 computational social science (CSS) benchmarks.

\begingroup
\begin{table*}[ht] 
\setlength{\tabcolsep}{0.8pt} 
\renewcommand{\arraystretch}{1.1} 
\centering
\begin{tabular}[t]{lcccccccc}
\toprule
\textbf{\footnotesize{Paper}} & 
$\begin{array}{c}
     \text{\textbf{\footnotesize{model}}}  \\
     \text{\textbf{\footnotesize{families}}}      
\end{array}$
& 
$\begin{array}{c}
     \text{\textbf{\footnotesize{\# of}}}  \\
     \text{\textbf{\footnotesize{model}}} \\  
     \text{\textbf{\footnotesize{versions}}}
\end{array}$  
& 
$\begin{array}{c}
     \text{\textbf{\footnotesize{\# of}}} \\
     \text{\textbf{\footnotesize{data-}}} \\
     \text{\textbf{\footnotesize{sets}}}        
\end{array}$
& 
$\begin{array}{c}
     \text{\textbf{\footnotesize{\# of}}}  \\
     \text{\textbf{\footnotesize{metrics}}}        
\end{array}$
& 
$\begin{array}{c}
     \text{\textbf{\footnotesize{Zero/}}}  \\
     \text{\textbf{\footnotesize{few}}}  \\
     \text{\textbf{\footnotesize{shot}}}  
\end{array}$
& 
$\begin{array}{c}
     \text{\textbf{\footnotesize{Lang.}}} 
\end{array}$
& 
$\begin{array}{c}
     \text{\textbf{\footnotesize{Dis-}}}  \\
     \text{\textbf{\footnotesize{agree.}}}  
\end{array}$
& 
$\begin{array}{c}
     \text{\textbf{\footnotesize{LLM as}}}  \\
     \text{\textbf{\footnotesize{Anno.}}}  
\end{array}$ \\
\midrule
$\begin{array}{l}
    \text{\citep{lee2023can}}
\end{array}$
& 
$\begin{array}{c}
     \text{{\footnotesize{GPT,Vicuna,}}}  \\
     \text{{\footnotesize{Flan,OPT-IML}}}  
\end{array}$
& 9 
& 6 & 4 & z\&f & {en} & \cmark & \xmark \\
\midrule
$\begin{array}{l}
    \text{\citep{santurkar2023whose}}
\end{array}$
& 
$\begin{array}{c}
     \text{{\footnotesize{GPT,Jurassic}}}  
\end{array}$
& 9 
& 1 & 3 & z & {en} & \cmark & \xmark \\
\midrule
$\begin{array}{l}
    \text{\citep{Ziems_Held_Shaikh_Chen_Zhang_Yang_2023}}
\end{array}$
& 
$\begin{array}{c}
     \text{{\footnotesize{GPT,Flan}}}  
\end{array}$
& 14 
& 20 & 2 & z\&f & {en} & \xmark & \cmark \\
\midrule
$\begin{array}{l}
    \text{\citep{zhu2023can}}
\end{array}$
& 
$\begin{array}{c}
     \text{{\footnotesize{GPT}}}  
\end{array}$
& 1 
& 5 & 5 & z\&f & {en} & \xmark & (\cmark) \\
\midrule
$\begin{array}{l}
    \text{\citep{gilardi2023chatgpt}}
\end{array}$
& 
$\begin{array}{c}
     \text{{\footnotesize{GPT}}}  
\end{array}$
& 1 
& 4 & 2 & z & {en+} & \xmark & \cmark \\
\midrule
$\begin{array}{l}
    \text{\citep{tornbergchatgpt}}
\end{array}$
& 
$\begin{array}{c}
     \text{{\footnotesize{GPT}}}  
\end{array}$
& 1 
& 1 & 3  & z & {en} & \xmark & \cmark \\
\midrule
$\begin{array}{l}
    \text{\citep{mohta2023large}}
\end{array}$
& 
$\begin{array}{c}
     \text{{\footnotesize{Vicuna,}}}  \\
     \text{{\footnotesize{Flan,Llama}}}  \\
\end{array}$
& 9 
& 5 & 3 & z & {en,fr,nl} & \xmark & \xmark \\
\midrule
$\begin{array}{l}
    \text{\citep{ding2022gpt}}
\end{array}$
& 
$\begin{array}{c}
     \text{{\footnotesize{GPT}}}  
\end{array}$
& 1 
& 4 & 4 & z\&f & {en+} & \xmark & \cmark \\
\midrule
$\begin{array}{l}
    \text{\citep{he2023annollm}}
\end{array}$
& 
$\begin{array}{c}
     \text{{\footnotesize{GPT}}}  
\end{array}$
& 1 
& 3 & 1 & z\&f & {en} & \xmark & \cmark \\
\midrule
$\begin{array}{l}
    \text{\citep{huang2023chatgpt}}
\end{array}$
& 
$\begin{array}{c}
     \text{{\footnotesize{GPT}}}  
\end{array}$
& 1 
& 1 & 2 & z & {en} & \xmark & \cmark \\
\midrule
$\begin{array}{l}
    \text{\citep{goel2023llms}}
\end{array}$
& 
$\begin{array}{c}
     \text{{\footnotesize{Palm}}}  
\end{array}$
& 1 
& 1 & 3 & f & {en} & \xmark & \cmark \\
\midrule
$\begin{array}{l}
    \text{\citep{wang2021want}}
\end{array}$
& 
$\begin{array}{c}
     \text{{\footnotesize{GPT}}}  
\end{array}$
& 1 
& 9 & 2 & f & {en} & \xmark & \cmark \\
\bottomrule
\end{tabular}
\caption{\label{citation-guide} Overview on LLM's as Annotators ({\footnotesize Language codes follow ISO 639, en+: predominantly English, with some additional language explorations}) }
\label{table:llm_anno_summary}
\end{table*}
\endgroup
\paragraph{Language: }The majority of these studies measure LLM performance on English corpora (see Table \ref{table:llm_anno_summary}). However, \citet{ding2022gpt} conduct tests to understand the possibility of using GPT on non-English corpora and \citet{mohta2023large} investigate the performance of open source LLMs on French, Dutch and English natural language inference (NLI) tasks. Thus far, models have shown better performance on English related tasks and performed notably poorly on low-resource languages \citet{srivastava2022beyond}. While \citet{ding2022gpt} see potential for GPT on languages other than English, \citet{mohta2023large} observe a considerable decline in performance with non-English languages.

\paragraph{Annotator Disagreement: } All studies referenced thus far assume the existence of a singular ground truth label for a given sample. 
\textcolor{black}{There has, however, been a shift in thinking across machine learning towards a collectivist approach, meaning the inclusion of all annotator perspectives rather than having a majority voted ground truth \citep{uma2021learning, Prabhakaran_Mostafazadeh_Davani_Diaz_2021, Cabitza_Campagner_Basile_2023, Rottger_Vidgen_Hovy_Pierrehumbert_2022, Nie_Zhou_Bansal_2020, Pavlick_Kwiatkowski_2019}. In this context, \citet{lee2023can} explore whether LLMs can capture the human opinion distribution. Additionally, \citet{santurkar2023whose} investigate the alignment between LLMs and human annotators with respect to the opinions and perspectives reflected in response to subjective questions.}
From Table \eqref{table:llm_anno_summary} we can see that the latter two studies which investigate the performance of LLMs on opinion distributions don't yet deem them ready as annotators. However, all studies that investigate the capabilities of GPT as an annotator within the traditional framework of majority voted labels agree with varying degrees that LLMs have the potential to disrupt the annotation process. Within this paradigm of majority voting, the sole exception to the consensus  
is expressed by \citet{mohta2023large} who conclude that LLMs have not yet attained a sufficient level for the annotation of datasets. Notably, amongst the cited studies, they are the sole study to only use open source LLMs and not consider best performing closed source alternatives (see Table \ref{table:llm_anno_summary}).
\paragraph{Models: }As mentioned in the previous paragraph, the predominant focus across all studies lies on models belonging to the GPT series. The remaining models under consideration are mostly open-source options, with Flan being the second most investigated, succeeded by Vicuna.
Table \ref{table:llm_anno_summary} highlights that only four studies explored model families beyond GPT. 
Notably, these same studies explored multiple versions of a given model (\textit{"\# of model versions"}). 
In contrast, the remaining studies exclusively assessed a singular model. More details on the exact versions can be found in table \ref{table:llm_anno_model_versions} (Appendix A).  

\paragraph{Temperature Parameter: }Not all studies mention the settings of their temperature parameter. However, both \citet{tornbergchatgpt, gilardi2023chatgpt} investigate the variability in responses by experimenting with lower (0.2) and high (1.0) temperature settings. They find that LLMs have higher consistency with lower temperatures without sacrifices in accuracy and thus recommend lower values for annotation tasks.
\citet{Ziems_Held_Shaikh_Chen_Zhang_Yang_2023} and \citet{goel2023llms} opt for a temperature of 0 throughout their study, aiming to ensure consistent and reproducible results across their LLM analysis.

\paragraph{Prompting: } \citet{wang2021want} and \citet{goel2023llms} investigate the efficacy of LLMs as annotators using only few-shot prompting. In contrast, five of the subsequent studies experiment with both zero- and few-shot prompting. Additionally, five other studies employ zero-shot prompting for their annotation tasks (see Table \ref{table:llm_anno_summary}). The outcomes of the experiments comparing zero-shot and few-shot prompting show inconsistency. \citet{mohta2023large} experience superior performance using few-shot prompting, while \citet{ding2022gpt} find that few-shot prompting does not yield superior results across all their approaches. \citet{he2023annollm} report a decrease in performance with few-shot prompting for their specific task. \citet{Ziems_Held_Shaikh_Chen_Zhang_Yang_2023} conclude that improvements from few-shot prompting are inconsistent across their experiments, suggesting that achieving more substantial gains would require increased efforts in refining the prompting process.

\begingroup
\begin{table}[hbt!]
\setlength{\tabcolsep}{4.pt} 
\renewcommand{\arraystretch}{1.1} 
\centering
\begin{tabular}{lllllll}
\rotatebox{0}{\textbf{\footnotesize{Paper}}} & 
    \rotatebox{90}{\textbf{\footnotesize{Accuracy}}} & 
    \rotatebox{90}{\textbf{\footnotesize{F1}}} & 
    \rotatebox{90}{\textbf{\footnotesize{Precision}}} & 
    \rotatebox{90}{\textbf{\footnotesize{Recall}}} & 
    \rotatebox{90}{\textbf{\footnotesize{Reliability}}} & 
    \rotatebox{90}{\textbf{\footnotesize{Other}}} \\
\toprule
    $\begin{array}{l}
        \text{\citep{lee2023can}}
    \end{array}$
    & {\footnotesize \cmark} & - & - & - & - & {\footnotesize \cmark} \\ \hline 
    
    $\begin{array}{l}
        \text{\citep{santurkar2023whose}}
    \end{array}$
    & - & - & - & - & - & {\footnotesize \cmark} \\ \hline

    $\begin{array}{l}
        \text{\citep{Ziems_Held_Shaikh_Chen_Zhang_Yang_2023}}
    \end{array}$
    & - & {\footnotesize \cmark} & - & - & {\footnotesize \cmark} & - \\ \hline
    
    $\begin{array}{l}
        \text{\citep{zhu2023can}}
    \end{array}$
    & {\footnotesize \cmark} & {\footnotesize \cmark} & {\footnotesize \cmark} & {\footnotesize \cmark} & - & {\footnotesize \cmark} \\ \hline
    
    $\begin{array}{l}
        \text{\citep{gilardi2023chatgpt}}
    \end{array}$
    & {\footnotesize \cmark} & - & - & - & {\footnotesize \cmark} & - \\ \hline

    $\begin{array}{l}
        \text{\citep{tornbergchatgpt}}
    \end{array}$
    & {\footnotesize \cmark} & - & - & - & {\footnotesize \cmark} & {\footnotesize \cmark} \\ \hline

    $\begin{array}{l}
        \text{\citep{mohta2023large}}
    \end{array}$
    & {\footnotesize \cmark} & {\footnotesize \cmark} & - & - & - & {\footnotesize \cmark} \\ \hline
    
    $\begin{array}{l}
        \text{\citep{ding2022gpt}}
    \end{array}$
    & {\footnotesize \cmark} & {\footnotesize \cmark} & {\footnotesize \cmark} & {\footnotesize \cmark} & - & - \\ \hline
    
    $\begin{array}{l}
        \text{\citep{he2023annollm}}
    \end{array}$
    & {\footnotesize \cmark} & - & - & - & - & - \\ \hline
    
    $\begin{array}{l}
        \text{\citep{huang2023chatgpt}}
    \end{array}$
    & {\footnotesize \cmark} & - & - & - & - & {\footnotesize \cmark} \\ \hline
    
    $\begin{array}{l}
        \textcolor{black}{\text{\citep{goel2023llms}}}
    \end{array}$
    & - & {\footnotesize \cmark} & {\footnotesize \cmark} & {\footnotesize \cmark} & - & {\footnotesize \cmark} \\ \hline

    $\begin{array}{l}
        \text{\citep{wang2021want}}
    \end{array}$
    & {\footnotesize \cmark} & - & - & - & - & {\footnotesize \cmark}
    \\
\bottomrule   
\end{tabular}
\caption{\label{citation-guide} Evaluation Metrics across Papers}
\label{table:llm_anno_eval}
\end{table}
\endgroup

\paragraph{Evaluation: }Nearly all studies assess their outcomes using metrics such as accuracy or F1. 
\citet{santurkar2023whose} deviate from these conventional performance metrics as, their primary focus lies in evaluating representation. This emphasis leads them to assess LLMs responses based on metrics measuring representativeness, steerability, and consistency \citep{santurkar2023whose}. 
In addition to accuracy and F1, three studies utilise metrics such as precision and recall, while three other studies employ different reliability measures to evaluate inter-coder agreement. \citet{tornbergchatgpt, santurkar2023whose} specifically investigate model bias, whereas \citet{huang2023chatgpt} evaluate the natural language explanations (NLE) that LLMs can provide for their predictions. For the evaluation of LLM and human opinion distributions, \citet{lee2023can} use entropy, Jensen-Shannon divergence (JSD), and the Human Distribution Calibration Error (DistCE) introduced by \citet{baan_stop_2022}.
Two studies have conducted error analyses. \citet{huang2023chatgpt} observe that the instances of disagreement, comprising 20\% in their study, align more closely with lay-people's perspectives. Similarly, \citet{Ziems_Held_Shaikh_Chen_Zhang_Yang_2023} conclude that in their error analysis, the LLM tends to default to more common label stereotypes.
Given the reported accuracy-based performance of LLMs on labelling tasks, it is important to broaden metrics to include more representational measures. For example, \citet{Ziems_Held_Shaikh_Chen_Zhang_Yang_2023} omit measuring bias in their study, concluding that larger, instruction-tuned models demonstrate superior performance. However, \citet{srivastava2022beyond} caution that larger models tend to amplify bias. 

\subsection{Benefits}
\label{subsection:benefits}
\label{subsection:benefits} 
\citet{tornbergchatgpt} finds that gpt-4 consistently surpasses the performance of both crowd-workers and expert coders, and the cost associated with labeling a sample is orders of magnitude lower for LLMs compared to humans. \citet{wang2021want} provide a detailed explanation that, in their experiments, utilising labels generated by the LLM resulted in a cost reduction ranging from 50\% to 96\%, while maintaining equivalent performance in downstream models. Similarly, \citet{goel2023llms} determine that the LLM reduces the total time of labelling by 58\% while maintaining a comparable baseline performance to medically trained annotators. \citet{gilardi2023chatgpt} demonstrate that the LLM shows superior quality compared to annotations obtained through Amazon Mechanical Turk (MTurk), while being approximately 30 times more cost-effective.
\citet{ding2022gpt} find that their approach attains nearly equivalent performance when labeling the same number of samples. However, when they double the amount of data labeled by the LLM, superior performance is achieved at only 10\% of the cost associated with human annotation \citep{ding2022gpt}.
LLMs not only entail lower costs than human annotators but also demonstrate significantly higher speeds in the labeling process \citep{tornbergchatgpt, wang2021want, ding2022gpt}.

In addition to diminished cost and time requirements, LLMs demonstrate the capability to provide explanations for their annotation \citep{mohta2023large}. \citet{huang2023chatgpt} find that ChatGPT generates explanations comparable, if not superior in clarity, to those produced by human annotators.

\subsection{Limitations}
\label{subsection:limitations}
As mentioned in Section \ref{sec:literature_review}, one limitation lies in the predominant development and testing of LLMs within the confines of the English language. An additional constraint associated with using LLMs as annotators is the challenge in formulating prompts and obtaining meaningful responses. Models might generate unconstrained responses \citep{goel2023llms} or might refrain from providing responses altogether as a result of the implementation of safeguarding measures. \citet{Ziems_Held_Shaikh_Chen_Zhang_Yang_2023} observed that models tended to predict beyond the presented labels and exhibited a tendency to abstain from responding to tasks deemed offensive. In the event that a model does provide a response, potential issues may arise in the form of bias. \citet{srivastava2022beyond} show that bias in LLMs increases in with scale and ambiguous contexts. \citet{santurkar2023whose} identify that LLMs demonstrate a singular perspective characterised by left-leaning tendencies. 
\citet{tornbergchatgpt} notes the absence of substantial disparities between expert annotators and LLMs, while underscoring the notable bias observed among annotators from MTurk. However, \citet{goel2023llms} underscore the importance of expert human annotators in attaining high-quality labels. 
\citet{lee2023can} express concerns regarding the population representation capabilities of current LLMs, whereas \citet{Ziems_Held_Shaikh_Chen_Zhang_Yang_2023} caution researchers to consider and mitigate the potential risks of bias in their applications through human-in-the-loop methods.

An additional noteworthy limitation in employing LLMs as annotators is their sensitivity to minor alterations in prompting \citep{loya2023exploring, sclar2023quantifying}. Both \citet{huang2023chatgpt} and \citet{Ziems_Held_Shaikh_Chen_Zhang_Yang_2023} assert the need for further research to comprehensively investigate the effects of prompting and determine optimal strategies for effective prompting. 
Lastly, it is important to note that these models show sub-optimal performance as annotators in 
tasks such as NLI, implicit hate classification, empathy or dialect detection \citep{lee2023can,Ziems_Held_Shaikh_Chen_Zhang_Yang_2023}.

\section{Results with the SEMEVAL 2023 Subjective Tasks Benchmark}
\label{sec:empirical_analysis}

As discussed above, most studies of LLMs as annotators still adopt a majority vote perspective, which is becoming increasingly questionable particularly for subjective tasks \cite{Akhtar_Basile_Patti_2021,Leonardelli_Menini_Palmero_Aprosio_Guerini_Tonelli_2021,uma2021learning,plank:EMNLP22:problem,Cabitza_Campagner_Basile_2023}.
We decided therefore to carry out a preliminary exploration of  the alignment between LLM and human judgment distributions on the  datasets used in the recent SEMEVAL 2023 Shared Task on Learning with Disagreement \cite{leonardelli_2023_semeval}.
Our analysis is centered on the extent to which the most frequently used model (GPT) matches human distribution on  datasets for  inherently subjective tasks. 
This was done  by extracting opinion distributions in the simplest and most straightforward manner possible: 
we directly prompt GPT to provide its estimation of the human opinion distribution and compare it against the baseline and optimal results from SemEval-2023.

\begingroup
\setlength{\tabcolsep}{1.2pt} 
\renewcommand{\arraystretch}{2.1} 
\begin{table}[ht]
\centering
\begin{tabular}[t]{llcccc}
\toprule
\arraycolsep=0.9pt\def\arraystretch{0.8}
$\begin{array}{c}
\textbf{\footnotesize Dataset} \end{array}$ 
& 
\arraycolsep=0.9pt\def\arraystretch{0.8}
$\begin{array}{c}
\textbf{{\footnotesize Task}}
\end{array}$
& 
\arraycolsep=0.9pt\def\arraystretch{0.8}
$\begin{array}{c}
\textbf{{\footnotesize Lang.}}
\end{array}$
&
\arraycolsep=0.9pt\def\arraystretch{0.5}
$\begin{array}{c}
     \text{\textbf{\footnotesize \# items}} \\     
     \text{{\scriptsize train}}  \\
     \text{{\scriptsize dev}}  \\
     \text{{\scriptsize test}}
\end{array}$  & 
\arraycolsep=0.9pt\def\arraystretch{0.8}
$\begin{array}{c}
    \text{\textbf{\footnotesize \% full}} \\
    \text{\textbf{\footnotesize agree.}}
\end{array}$ \\
\midrule
{\footnotesize MD-Agree.} &
    \arraycolsep=0.6pt\def\arraystretch{0.5} 
    \begin{tabular}[c]{@{}l@{}}
        \scriptsize {Offensiveness} \\ 
        \scriptsize {detection}
    \end{tabular} 
&
{\footnotesize en} &
\arraycolsep=0.6pt\def\arraystretch{0.6} 
$\begin{array}{c} 
    \text{{\scriptsize 6592}}  \\
    \text{{\scriptsize 1104}}  \\ 
    \text{{\scriptsize 3057}} 
\end{array}$
& {\footnotesize 42\%} \\
{\footnotesize HS-Brexit} & 
    \arraycolsep=0.6pt\def\arraystretch{0.5} 
    \begin{tabular}[c]{@{}l@{}}
        \scriptsize {Offensiveness} \\ 
        \scriptsize {detection}
    \end{tabular} &
{\footnotesize en} &
\arraycolsep=0.6pt\def\arraystretch{0.6} 
$\begin{array}{c}  
    \text{{\scriptsize 784 }}  \\  
    \text{{\scriptsize 168 }}  \\ 
    \text{{\scriptsize 168 }}  
\end{array}$
& {\footnotesize 69\%} \\
{\footnotesize ConvAbuse} &
\arraycolsep=0.6pt\def\arraystretch{0.5} 
    \begin{tabular}[c]{@{}l@{}}
        \scriptsize {Abusiveness} \\ 
        \scriptsize {detection}
    \end{tabular}  &
{\footnotesize en} &
\arraycolsep=0.6pt\def\arraystretch{0.6} 
$\begin{array}{c} 
    \text{{\scriptsize  2398}}  \\  
    \text{{\scriptsize 812}}  \\ 
    \text{{\scriptsize 840}}  
\end{array}$
& {\footnotesize 86\%} \\
{\footnotesize ArMIS} &
    \arraycolsep=0.6pt\def\arraystretch{0.5} 
    \begin{tabular}[c]{@{}l@{}}
        \scriptsize {Misogyny} \\ 
        \scriptsize {and sexism} \\
        \scriptsize {detection}
    \end{tabular} &
{\footnotesize ar} &
\arraycolsep=0.6pt\def\arraystretch{0.6} 
$\begin{array}{c} 
    \text{{\scriptsize 657}}  \\  
    \text{{\scriptsize 141}}  \\ 
    \text{{\scriptsize 145}}  
\end{array}$
& {\footnotesize 65\%} \\
\bottomrule
\end{tabular}
\caption{Dataset statistics \citep{leonardelli_2023_semeval} ({\footnotesize Language codes follow ISO 639})}
\label{table:datasets}
\end{table}
\endgroup

\subsection{Datasets} 
\label{section:datasets_sub}
We leverage four datasets from SemEval2023 on "Learning with Disagreements" for the empirical analysis. All four datasets focus on subjective tasks and  contain human annotated target distributions 
that we compare to the LLM predictions. Table \ref{table:datasets} contains key statistics on the datasets \citep{leonardelli_2023_semeval}.
\paragraph{Multi-Domain Agreement:} 
MD-Agreement \citep{Leonardelli_Menini_Palmero_Aprosio_Guerini_Tonelli_2021} 
is the dataset with the lowest amount of annotator agreement amongst these subjective tasks. Each example was labelled by 5 annotators and was created using English tweets from three domains (BLM, Election2020 and Covid-19). 
\paragraph{Hate Speech on Brexit:} HS-Brexit \citep{Akhtar_Basile_Patti_2021} 
was  constructed from English tweets using keywords related to immigration and Brexit. Each example was labelled by 6 annotators with 69\% of items having total annotator agreement. 
\paragraph{ConvAbuse: } 
ConvAbuse \citep{curry2021convabuse} consists of English conversational text collected from dialogue between users and two conversational AI systems. 
Each example was labelled by between 3 and 8 annotators. 86\% of items have total annotator agreement.
\paragraph{Arabic Misogyny and Sexism: } ArMIS \citep{Almanea_Poesio_2022} is the only non-English language task and serves to study the effect on sexism judgements particularly with respect to the annotators leanings towards conservatism or liberalism. Each example was labelled by 3 annotators with 65\% of items having total annotator agreement. 


\subsection{Experimental Parameters}
\label{subsection:experiment_params}
We explore the capability of \texttt{gpt-3.5-turbo} to generate opinion distributions for the test data of each SemEval2023 task.
Given the sensitivity of LLMs to minor changes in input \citep{loya2023exploring, sclar2023quantifying}, we maintain a uniform prompt structure across various tasks and let the LLM assume the role of an expert annotator who considers multiple worldviews and cultural nuances. Modifications are made only on the words related to the respective task under consideration. For instance, in the case of HS-Brexit, the LLM specialises in "hate speech detection," whereas in the ConvAbuse dataset its specialisation lies in "abusiveness detection."
ArMIS is approached with slight variation due to the presence of Arabic text. In this instance, we explore two approaches: one involves prompting the models in English and providing them with the Arabic text that requires labelling, while the second approach uses an Arabic prompt (a translated version of the English prompt).

As mentioned in Section \ref{sec:literature_review} there is some variability both among and within studies regarding the preferred prompting approach for LLM annotation. However, given that the multiple studies indicate limited benefits from few-shot prompting, we opt for zero-shot prompting in our tasks. The expectation of a model's output on a labelling task is to be consistent. In order to achieve such consistent and reproducible results we set the temperature parameter across our models to zero such as \citet{Ziems_Held_Shaikh_Chen_Zhang_Yang_2023}. \citet{gilardi2023chatgpt} suggest that a lower temperature value might be preferable for annotation task as it increases consistency without decreasing accuracy across their empirical analysis. 

\subsection{Evaluation Metrics}
\label{subsection:eval_metrics}
We compare the performance of GPT to both the Semeval2023 baseline model as well as the top-performing model on each task. \citet{leonardelli_2023_semeval} evaluate point predictions using the F1 measure \eqref{eq:f1} and distribution similarity using Cross-Entropy (CE) \eqref{eq:CE}. To ensure comparability we use both of these in our analysis. 

\begin{equation}\label{eq:f1}
    {\footnotesize \text{F1} = \frac{2*TP}{2*TP+FP+FN}}
\end{equation}
\begin{equation}\label{eq:CE}
    {\footnotesize \text{CE}(y_n,\hat{y}_n) = -\sum_{n=1}^N y_n \log (\hat{y}_n)} \, ,
\end{equation}
where $y_{n}$ is a sample opinion distribution annotated by humans and $\hat{y}_n$ the LLMs predicted distribution for that sample.
In addition to the above, we also use Shannon's entropy 
to visualise human and LLM uncertainties. 

\begingroup
\setlength{\tabcolsep}{0.8pt}
\renewcommand{\arraystretch}{1.2}
\begin{table*}[hbt!]
\centering
\begin{tabular}[t]{l|ccc|ccc|ccc|cccc}
\toprule
    \multirow{2}{*}{\footnotesize{}} 
    & \multicolumn{3}{c}{\footnotesize{MD-Agree.}} \vline 
    & \multicolumn{3}{c}{\footnotesize{HS-Brexit}} \vline 
    & \multicolumn{3}{c}{\footnotesize{ConvAbuse}} \vline 
    & \multicolumn{4}{c}{\footnotesize{ArMIS}} 
    \\
    \midrule
    & \footnotesize{gpt} 
    & \arraycolsep=0.6pt\def\arraystretch{0.5} $\begin{array}{c} \text{{\footnotesize SE}}  \\  \text{\scriptsize{(baseline)}} \end{array}$
    & $\arraycolsep=0.6pt\def\arraystretch{0.5} \begin{array}{c} \text{{\footnotesize SE}}  \\  \text{\scriptsize{(best)}} \end{array}$
    & \footnotesize{gpt} 
    & $\arraycolsep=0.6pt\def\arraystretch{0.5} \begin{array}{c} \text{{\footnotesize SE}}  \\  \text{\scriptsize{(baseline)}} \end{array}$
    & $\arraycolsep=0.6pt\def\arraystretch{0.5} \begin{array}{c} \text{{\footnotesize SE}}  \\  \text{\scriptsize{(best)}} \end{array} $
    & \footnotesize{gpt}
    & \arraycolsep=0.6pt\def\arraystretch{0.5} $\begin{array}{c} \text{{\footnotesize SE}}  \\  \text{\scriptsize{(baseline)}} \end{array}$
    & \arraycolsep=0.6pt\def\arraystretch{0.5} $\begin{array}{c} \text{{\footnotesize SE}}  \\  \text{\scriptsize{(best)}} \end{array}$
    & \arraycolsep=0.6pt\def\arraystretch{0.5} $\begin{array}{c} \text{{\footnotesize gpt}}  \\  \text{\scriptsize{(english)}} \end{array}$
    & \arraycolsep=0.6pt\def\arraystretch{0.5} $\begin{array}{c} \text{{\footnotesize gpt}}  \\  \text{\scriptsize{(arabic)}} \end{array}$
    & \arraycolsep=0.6pt\def\arraystretch{0.5} $\begin{array}{c} \text{{\footnotesize SE}}  \\  \text{\scriptsize{(baseline)}} \end{array} $
    & \arraycolsep=0.6pt\def\arraystretch{0.5} $\begin{array}{c} \text{{\footnotesize SE}}  \\  \text{\scriptsize{(best)}} \end{array}$
    \\
    
    \midrule
    {\footnotesize $F1 \uparrow$}     
    & \footnotesize{0.520} & \textbf{\footnotesize 0.534} &\textit{\footnotesize 0.846} 
    & \footnotesize{0.696} & \textbf{\footnotesize 0.842} & \textit{\footnotesize 0.929} 
    & \textbf{\footnotesize 0.902} & \footnotesize{0.741} & \textit{\footnotesize 0.942}  
    & \textbf{\footnotesize 0.448} & {\footnotesize 0.256}& \footnotesize{0.417} & \textit{\footnotesize 0.832}  
    \\
        
    {\footnotesize $CE \downarrow$}
    & \textbf{\footnotesize 3.829} & \footnotesize{7.385} & \textit{\footnotesize 0.472} 
    & \footnotesize{5.037} & \textbf{\footnotesize 2.715} & \textit{\footnotesize 0.235} 
    & \footnotesize{3.746} & \textbf{\footnotesize 3.484} & \textit{\footnotesize 0.185}  
    & \textbf{\footnotesize 5.828} & \footnotesize{6.667}& \footnotesize{8.908} & \textit{\footnotesize 0.469} 
    \\
    
    
    
\bottomrule
\end{tabular}
\caption{\label{table:empirical_results} Prompting gpt-3.5-turbo directly vs. baselline \& best results from SemEval2023 (SE)}
\end{table*}
\endgroup

\begin{figure*}[hbt!]
  \centering {\includegraphics[width=0.95\textwidth]{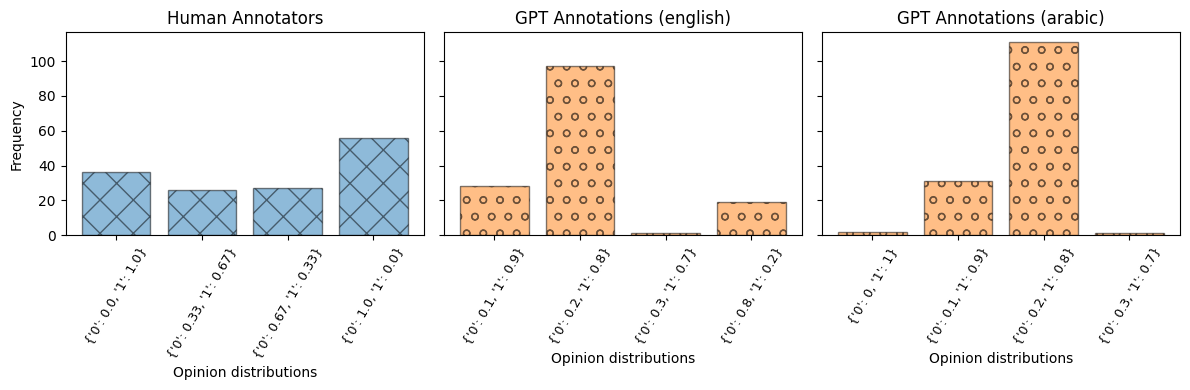}} \vspace{-0.45cm}
  \caption{ArMIS opinion distributions}
  \label{figure:hist_armis_distribution}
\end{figure*}

\begin{figure}[hbt!]
  \centering {\includegraphics[width=0.43\textwidth]{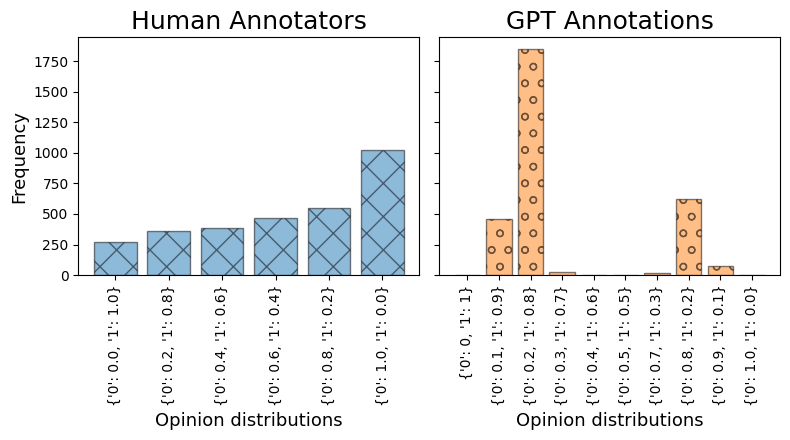}}\vspace{-0.38cm}
  \caption{MD-Agreement opinion distributions}
  \label{figure:hist_md_distribution}
\end{figure}

\begin{figure}[hbt!]
  \centering {\includegraphics[width=0.45\textwidth]{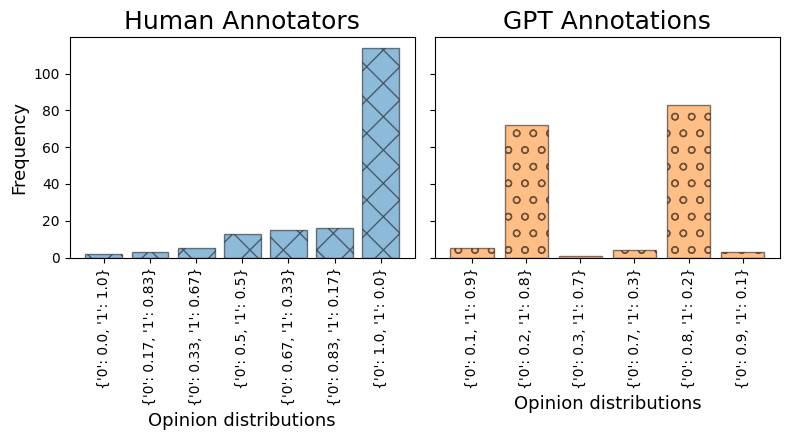}} \vspace{-0.38cm}
  \caption{HS-Brexit opinion distributions}
  \label{figure:hist_hs_brexit_distribution}
\end{figure}

\begin{figure}[hbt!]
  \centering {\includegraphics[width=0.45\textwidth]{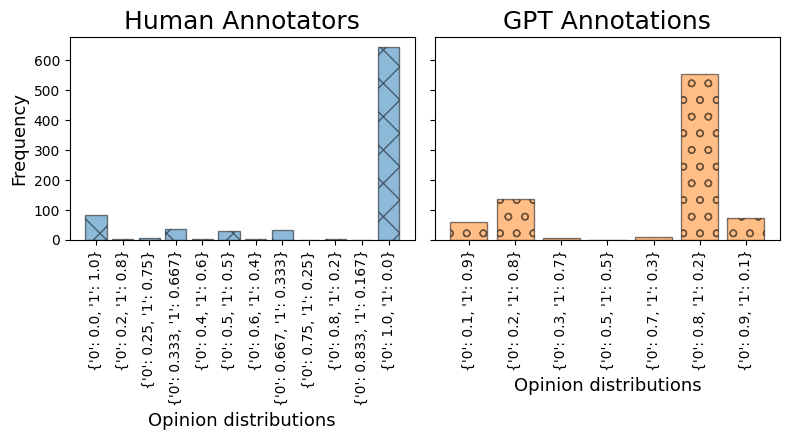}}\vspace{-0.38cm}
  \caption{ConvAbuse opinion distributions}
  \label{figure:hist_convabuse_distribution}
\end{figure}

\begingroup
\setlength{\tabcolsep}{12pt} 
\renewcommand{\arraystretch}{1.} 
\begin{table}[hbt!]
\centering
\begin{tabular}{lcc}
    \multirow{2}{*}{\footnotesize{}} 
    & \multicolumn{2}{c}{\footnotesize{\textbf{Categorisation of Errors}}}
    \\
\toprule
    \footnotesize{\textbf{Dataset}}
    & \footnotesize{\textbf{FP}} 
    & \footnotesize{\textbf{FN}} 
    \\
\midrule
    \footnotesize MD-Agree & 
    \footnotesize 96.87\% &
    \footnotesize 3.13\% 
    \\ 
    \hline
    
    \footnotesize HS-Brexit & 
    \footnotesize 100.00\% & 
    \footnotesize 0.00\% \\ 
    \hline
    
    \footnotesize ConvAbuse & 
    \footnotesize 91.11\% & 
    \footnotesize 8.89\% \\ 
    \hline
     
    \footnotesize ArMIS (english) & 
    \footnotesize 95.71\% & 
    \footnotesize 4.29\% \\ 
    \hline
    
    \footnotesize ArMIS (arabic) &
    \footnotesize 100.00\% & 
    \footnotesize 0.00\% \\ 
    \bottomrule
\end{tabular}
\caption{{Categorisation of errors into percentage that are False Positive vs. False Negative.} \textit{GPT 3.5-turbo} across different SemEval2023 tasks }
\label{table:false_positive_false_negative}  
\end{table}
\endgroup

\subsection{Results}
\label{subsec:empirical_results}
Figures \ref{figure:hist_armis_distribution},\ref{figure:hist_md_distribution}, \ref{figure:hist_hs_brexit_distribution} and \ref{figure:hist_convabuse_distribution} contrast the frequency of opinion distributions of human annotators with those predicted by GPT for each SemEval task.
We observe that when prompted directly for opinion distributions, the model shows a tendency towards bimodal predictions, with a notable preference for the following opinion distributions: \texttt{\{"0":0.2, "1":0.8\}} and \texttt{\{"0":0.8,"1":0.2\}}.

Another notable observation is evident in Figure \ref{figure:hist_armis_distribution}, where we observe a bias towards assigning greater weight to the sexist class ('1') when prompting the LLM with Arabic text.
In fact, when these distributions are simplified to a majority-based label, all test samples are categorised as sexist, a pattern not observed when the LLM was prompted with English text. The difference is also evident in the F1 performance (Table \ref{table:empirical_results}). 
The LLM prompted in Arabic only achieves an F1 score of 0.256, whereas prompting the LLM in English results in a score of 0.448, suggesting that LLMs perhaps understand the English prompt 
better than the Arabic one. 
The overall performance, however, remains significantly lower compared to other datasets, both in terms of F1 and CE metrics. This finding aligns with \citet{mohta2023large} who find that LLMs perform better on English 
datasets.

Table \ref{table:empirical_results} highlights that while the simplistic baseline performance can be matched
, it consistently falls short of the performance achieved by a specifically fine-tuned model on both F1 and CE scores (SE best).

\begin{figure*}[hbt!]
  \centering {\includegraphics[width=0.98\textwidth]{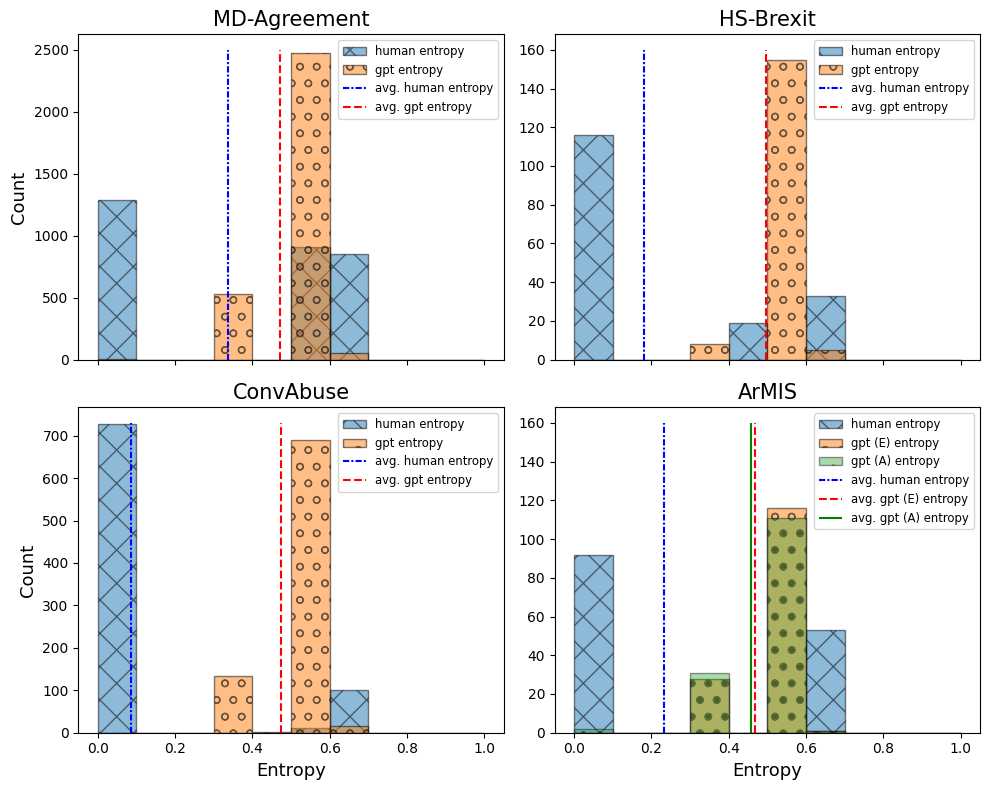}}
  \caption{Histogram showing human and GPT entropy}
  \label{figure:Entropy_all}
\end{figure*}

{A further examination of the errors when using the final majority voted labels 
reveals a higher tendency for false positive errors (see Table \ref{table:false_positive_false_negative}). This indicates that  
GPT is biased towards annotating samples as offensive, abusive, and misogynistic.}
Prompting the LLM to directly return opinion distributions results in higher average entropy values across all four datasets when compared to the average human entropy values (Figure \ref{figure:Entropy_all}). This stems from the observations made in the initial four figures. With the exception of the Arabic prompt, GPT consistently provides opinion distributions that allocate a small proportion to both classes rather than assigning 100 percent to one class. This leads to increased per sample entropy and thereby overall higher average entropy.




\section{Conclusion}
\label{sec:conclusions}
The overview section is not intended to provide an exhaustive review; however, the variety of tasks, datasets and approaches within the surveyed papers offers first insight into the efficacy of using LLMs to annotate data.
Despite the mentioned limitations, 
the overall findings show a degree of consensus and positive outlook towards the use of LLMs as data annotators within the majority voting 
paradigm.

Our initial observations suggest that, when directly prompted, GPT tends to produce label distributions that are not strongly aligned with human opinion distributions.
Furthermore, also consistent with prior research, the LLM shows superior performance on English language tasks compared to non-English text, while also showing potential bias in its responses.
However, given that LLMs are trained to predict next tokens,  
directly obtaining opinion distributions from them has inherent limitations. 
Hence, in future work, we aim to explore further approaches to extracting the probability distributions such as through normalising the log probabilities \citep{santurkar2023whose} or through Monte Carlo estimation \citep{lee2023can}.

\section*{Ethical statement}
Our study exclusively used pre-existing datasets for experimentation purposes. While the datasets contain instances of offensive language, 
our approach involved handling this content without direct human involvement. 

\section*{Acknowledgments}
Maja Pavlovic is supported by a Deep Mind PhD studentship to Queen Mary University. 
The work of Massimo Poesio is supported in part by the AINED Fellowship Grant \textit{Dealing with Meaning Variation in NLP}, NGF.1607.22.002. 


\nocite{*}
\section{Bibliographical References}\label{sec:reference}

\bibliographystyle{lrec-coling2024-natbib}
\bibliography{lrec-coling2024-example}


\newpage
\section*{Appendix A - Additional tables}
\label{appendix_tables}

\begingroup
\begin{table}[hbt!] 
\setlength{\tabcolsep}{0.5pt} 
\renewcommand{\arraystretch}{0.80} 
\centering
\begin{tabular}[t]{ll}
\toprule
    \textbf{\footnotesize{Paper}} & 
    $\begin{array}{c}
         \text{\textbf{\footnotesize{Datasets}}}      
    \end{array}$ \\
\midrule
    $\begin{array}{l}
        \text{\footnotesize\citep{lee2023can}}
    \end{array}$
    & 
    $\begin{array}{l}
       \text{\footnotesize{ANLI-R3, QNLI,}} \\
       \text{\footnotesize{ChaosNLI, PK2019}} \\
    \end{array}$ \\
\midrule
    $\begin{array}{l}
        \text{\footnotesize\citep{santurkar2023whose}}
    \end{array}$
    & 
    $\begin{array}{l}
       \text{\footnotesize{OpinionQA}}
    \end{array}$ \\
\midrule
    $\begin{array}{l}
        \text{\footnotesize\citep{Ziems_Held_Shaikh_Chen_Zhang_Yang_2023}}
    \end{array}$
    &
    $\begin{array}{l}
       \text{\footnotesize{Indian English dialect}} \\
       \text{\footnotesize{feature detection, Twitter}} \\
       \text{\footnotesize{Emotion detection, FLUTE,}} \\
       \text{\footnotesize{Latent Hatred, Reddit/}} \\
       \text{\footnotesize{Kaggle Humor data,}} \\
       \text{\footnotesize{Ideological Books Corpus,}} \\
       \text{\footnotesize{Misinfo Reaction Frames}} \\ 
       \text{\footnotesize{Corpus, Random Acts of }} \\ 
       \text{\footnotesize{Pizza, Semeval2016}} \\ 
       \text{\footnotesize{Stance Dataset,}} \\ 
      \text{\footnotesize{Temporal Word-in-}} \\ 
      \text{\footnotesize{Context benchmark,}} \\ 
      \text{\footnotesize{Coarse Discourse}} \\ 
      \text{\footnotesize{Sequence Corpus,}} \\ 
      \text{\footnotesize{TalkLife dataset, }} \\
      \text{\footnotesize{Winning Arguments }} \\
      \text{\footnotesize{Corpus, Wikipedia Talk}} \\
      \text{\footnotesize{Pages dataset,}} \\
      \text{\footnotesize{Conversations Gone}} \\ 
      \text{\footnotesize{Awry Corpus, Stanford}} \\
      \text{\footnotesize{Politeness Corpus,}}\\
      \text{\footnotesize{Hippocorpus, WikiEvents}}\\
      \text{\footnotesize{Article Bias Corpus,}} \\
      \text{\footnotesize{CMU Movie corpus dataset}} 
    \end{array}$ \\
\midrule
    $\begin{array}{l}
        \text{\footnotesize\citep{zhu2023can}}
    \end{array}$
    & 
    $\begin{array}{l}
       \text{\footnotesize{Stance Detection, Hate}} \\
       \text{\footnotesize{Speech, Sentiment}} \\
       \text{\footnotesize{Analysis, Bot Detection,}} \\
       \text{\footnotesize{Russo-Ukrainian Sentiment}}
    \end{array}$ \\
\midrule
    $\begin{array}{l}
        \text{\footnotesize\citep{gilardi2023chatgpt}}
    \end{array}$
    &
    $\begin{array}{l}
       \text{\footnotesize{Twitter Content moderation,}} \\
       \text{\footnotesize{US Congress, Newspaper}} \\
       \text{\footnotesize{article content moderation}}
    \end{array}$ \\
\midrule
    $\begin{array}{l}
        \text{\footnotesize\citep{tornbergchatgpt}}
    \end{array}$
    &
    $\begin{array}{l}
       \text{\footnotesize{Twitter Parliamentarian}} \\
       \text{\footnotesize{Database}}
    \end{array}$ \\
\midrule
    $\begin{array}{l}
        \text{\footnotesize\citep{mohta2023large}}
    \end{array}$
    &
    $\begin{array}{l}
       \text{\footnotesize{MM-IMDB, XNLI,}} \\
       \text{\footnotesize{Hateful memes,}} \\
       \text{\footnotesize{2 proprietary datasets}} \\
    \end{array}$ \\
\midrule
    $\begin{array}{l}
        \text{\footnotesize\citep{ding2022gpt}}
    \end{array}$
    & 
    $\begin{array}{l}
       \text{\footnotesize{SST2, CrossNER,}} \\
       \text{\footnotesize{FewRel, ASTEData-V2}}
    \end{array}$ \\
\midrule
    $\begin{array}{l}
        \text{\footnotesize\citep{he2023annollm}}
    \end{array}$
    &
    $\begin{array}{l}
       \text{\footnotesize{QK \textit{(user query \& keyword}}} \\
       \text{\footnotesize{relevance assessment), }} \\
       \text{\footnotesize{Word-inContext WiC,}} \\
       \text{\footnotesize{BoolQ}} 
    \end{array}$ \\
\midrule
    $\begin{array}{l}
        \text{\footnotesize\citep{huang2023chatgpt}}
    \end{array}$
    &
    $\begin{array}{l}
       \text{\footnotesize{LatentHatred}}
    \end{array}$ \\
\midrule
    $\begin{array}{l}
        \text{\footnotesize\citep{goel2023llms}}
    \end{array}$
    &
    $\begin{array}{l}
       \text{\footnotesize{Mimic-iv-note}}
    \end{array}$ \\
\midrule
    $\begin{array}{l}
        \text{\footnotesize\citep{wang2021want}}
    \end{array}$
    &
    $\begin{array}{l}
       \text{\footnotesize{XSum, Gigaword,}} \\
       \text{\footnotesize{SQuAD, SST-2,}} \\
       \text{\footnotesize{CB TREC, AGNews, }} \\
       \text{\footnotesize{DBPedia, RTE}} \\
    \end{array}$ \\
\bottomrule
\end{tabular}
\caption{\label{citation-guide} Datasets used across different studies }
\label{table:llm_anno_datasets}
\end{table}
\endgroup


\begingroup
\begin{table}[hbt!] 
\setlength{\tabcolsep}{0.5pt} 
\renewcommand{\arraystretch}{0.85} 
\centering
\begin{tabular}[t]{ll}
\toprule
    \textbf{\footnotesize{Paper}} & 
    $\begin{array}{l}
         \text{\textbf{\footnotesize{Model Versions}}}      
    \end{array}$ \\
\midrule
    $\begin{array}{l}
        \text{\footnotesize\citep{lee2023can}}
    \end{array}$
    &
    $\begin{array}{l}
       \text{\footnotesize{GPT (text-davinci-002\&;}} \\
       \text{\footnotesize{003); FlanT5 (large,xl,xxl),}} \\
       \text{\footnotesize{Flan UL2; Stable Vicuna;}} \\
       \text{\footnotesize{OPT-IML-M-S(1.3B)\&}} \\       
       \text{\footnotesize{(30B)}}       
    \end{array}$ \\
\midrule
    $\begin{array}{l}
        \text{\footnotesize\citep{santurkar2023whose}}
    \end{array}$
    &
    $\begin{array}{l}
       \text{\footnotesize{GPT(ada,davinci,}} \\
       \text{\footnotesize{text-ada-001,text-davinci-}} \\
       \text{\footnotesize{001\&002\&003); Jurassic}} \\
       \text{\footnotesize{(j1-Grande, j1-Jumbo,}} \\
       \text{\footnotesize{j1-Grande-v2 beta) }}
    \end{array}$ \\
\midrule
    $\begin{array}{l}
        \text{\footnotesize\citep{Ziems_Held_Shaikh_Chen_Zhang_Yang_2023}}
    \end{array}$
    &
    $\begin{array}{l}
       \text{\footnotesize{GPT (text-ada-001,}} \\ 
       \text{\footnotesize{text-babbage-001,}} \\ 
       \text{\footnotesize{text-curie-001, text-}} \\ 
       \text{\footnotesize{davinci-001\&002\&003,}} \\ 
       \text{\footnotesize{gpt-3.5-turbo, gpt-4);}} \\ 
       \text{\footnotesize{FlanT5 (small, base}} \\
       \text{\footnotesize{large, xl, xxl), Flan UL2}}

    \end{array}$ \\
\midrule
    $\begin{array}{l}
        \text{\footnotesize\citep{zhu2023can}}
    \end{array}$
    &
    $\begin{array}{l}
       \text{\footnotesize{gpt-3.5-turbo}}
    \end{array}$ \\
\midrule
    $\begin{array}{l}
        \text{\footnotesize\citep{gilardi2023chatgpt}}
    \end{array}$
    &
    $\begin{array}{l}
       \text{\footnotesize{gpt-3.5-turbo}}
    \end{array}$ \\
\midrule
    $\begin{array}{l}
        \text{\footnotesize\citep{tornbergchatgpt}}
    \end{array}$
    &
    $\begin{array}{l}
       \text{\footnotesize{gpt-4}}
    \end{array}$ \\
\midrule
    $\begin{array}{l}
        \text{\footnotesize\citep{mohta2023large}}
    \end{array}$
    &
    $\begin{array}{l}
       \text{\footnotesize{Instruct-BLIP-Flan-T5;}} \\
       \text{\footnotesize{Vicuna(13b v1.3\&v1.5),}} \\
       \text{\footnotesize{Vicuna (7b v1.5)}} \\
       \text{\footnotesize{InstructBLIP (7b \& 13b),}} \\
       \text{\footnotesize{Llama (7b, 13b),}} \\
       \text{\footnotesize{OpenLlama (13b)}} 
    \end{array}$ \\
\midrule
    $\begin{array}{l}
        \text{\footnotesize\citep{ding2022gpt}}
    \end{array}$
    &
    $\begin{array}{l}
       \text{\footnotesize{text-davinci-003}}
    \end{array}$ \\
\midrule
    $\begin{array}{l}
        \text{\footnotesize\citep{he2023annollm}}
    \end{array}$
    &
    $\begin{array}{l}
       \text{\footnotesize{gpt-3.5}}
    \end{array}$ \\
\midrule
    $\begin{array}{l}
        \text{\footnotesize\citep{huang2023chatgpt}}
    \end{array}$
    &
    $\begin{array}{l}
       \text{\footnotesize{ChatGPT}}
    \end{array}$ \\
\midrule
    $\begin{array}{l}
        \text{\footnotesize\citep{goel2023llms}}
    \end{array}$
    &
    $\begin{array}{l}
       \text{\footnotesize{PaLM 2}}
    \end{array}$ \\
\midrule
    $\begin{array}{l}
        \text{\footnotesize\citep{wang2021want}}
    \end{array}$
    &
    $\begin{array}{l}
       \text{\footnotesize{text-davinci}}
    \end{array}$ \\
\bottomrule
\end{tabular}
\caption{\label{citation-guide} Model versions used across studies }
\label{table:llm_anno_model_versions}
\end{table}
\endgroup


\newpage

\section*{Appendix B - Prompts} \label{appendix_prompts}
\paragraph{Multi-Domain Agreement: } \textit{You are an expert annotator in "offensiveness detection", who considers multiple worldviews and cultural nuances. Now assign probabilities for the text below, which is delimited by three dashes (-), for "offensive speech" or "not offensive speech". Return your probabilities in a json format with the key being 1 for 'offensive' and 0 for 'not offensive'. You obviously know that probabilities need to sum to 1.}

\paragraph{Hate Speech on Brexit: } \textit{You are an expert annotator in "hate speech detection", who considers multiple worldviews and cultural nuances. Now assign probabilities for the text below, which is delimited by three dashes (-), for "hateful speech" or "no hateful speech". Return your probabilities in a json format with the key being 1 for 'hate speech' and 0 for 'no hate speech'. You obviously know that probabilities need to sum to 1.}

\paragraph{ConvAbuse: } \textit{You are an expert annotator in "abusiveness detection", who considers multiple worldviews and cultural nuances. Now assign probabilities for the text below, which is delimited by three dashes (-), for "abusive speech" or "no abusive speech". Return your probabilities in a json format with the key being 1 for 'abusive speech' and 0 for 'no abusive speech'. You obviously know that probabilities need to sum to 1.}

\paragraph{Arabic Misogyny and Sexism: } \textit{You are an expert annotator in "sexism detection", who considers multiple worldviews and cultural nuances. Now assign probabilities for the text below, which is delimited by three dashes (-), for "sexist speech" or "no sexist speech". Return your probabilities in a json format with the key being 1 for 'sexist speech' and 0 for 'no sexist speech'. You obviously know that probabilities need to sum to 1.}\\ \\
Arabic prompt: 
\begin{figure}[hbt!] \centering {\includegraphics[width=0.51\textwidth]{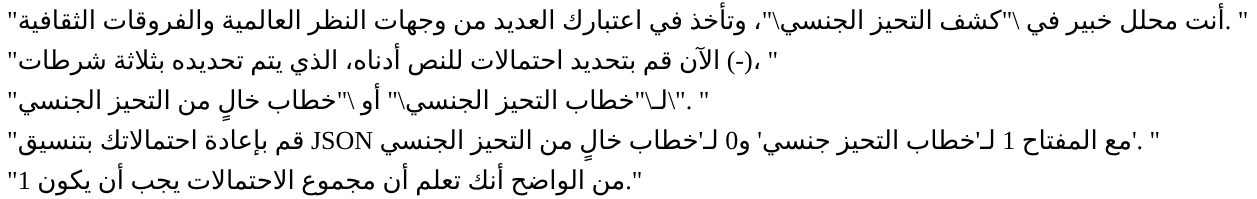}}
\end{figure}

\end{document}